\documentclass{article}
\pdfoutput=1
\usepackage[numbers]{natbib}

\usepackage[preprint]{neurips_2023}

\usepackage[utf8]{inputenc} % allow utf-8 input
\usepackage[T1]{fontenc}    % use 8-bit T1 fonts
\usepackage{hyperref}       % hyperlinks
\usepackage{url}            % simple URL typesetting
\usepackage{booktabs}       % professional-quality tables
\usepackage{amsfonts}       % blackboard math symbols
\usepackage{nicefrac}       % compact symbols for 1/2, etc.
\usepackage{microtype}      % microtypography
\usepackage{xcolor}         % colors
\usepackage{multicol}
\usepackage{multirow}

\usepackage{graphicx}
\graphicspath{{images/}}

\usepackage{caption}
\usepackage{subcaption}
\usepackage{wrapfig}
\usepackage{amsmath}

\usepackage{placeins}

\title{ The Forward-Forward Algorithm: \\ Characterizing Training Behavior}

\author{%
  Reece Adamson\\
  University of Massachusetts Amherst
}

\begin{document}

\maketitle

\begin{abstract}

  The Forward-Forward algorithm is an alternative learning method which consists of two forward passes rather than a forward and backward pass employed by backpropagation. Forward-Forward networks employ layer local loss functions which are optimized based on the layer activation for each forward pass rather than a single global objective function. This work explores the dynamics of model and layer accuracy changes in Forward-Forward networks as training progresses in pursuit of a mechanistic understanding of their internal behavior. Treatments to various system characteristics are applied to investigate changes in layer and overall model accuracy as training progresses, how accuracy is impacted by layer depth, and how strongly individual layer accuracy is correlated with overall model accuracy. The empirical results presented suggest that layers deeper within Forward-Forward networks experience a delay in accuracy improvement relative to shallower layers and that shallower layer accuracy is strongly correlated with overall model accuracy.
\end{abstract}

% System - What carries out the computation that produces that phenomenon?
% Task - What was the system designed or evolved to do?
% Environment - What is the context in which the system does the task?
% Phenomenon - What computation is being studied

\section{Introduction}

\begin{wrapfigure}{r}{0.3\textwidth}
\label{fig:ff-simple}
  \vspace{-10pt}
  \includegraphics[width=0.3\textwidth]{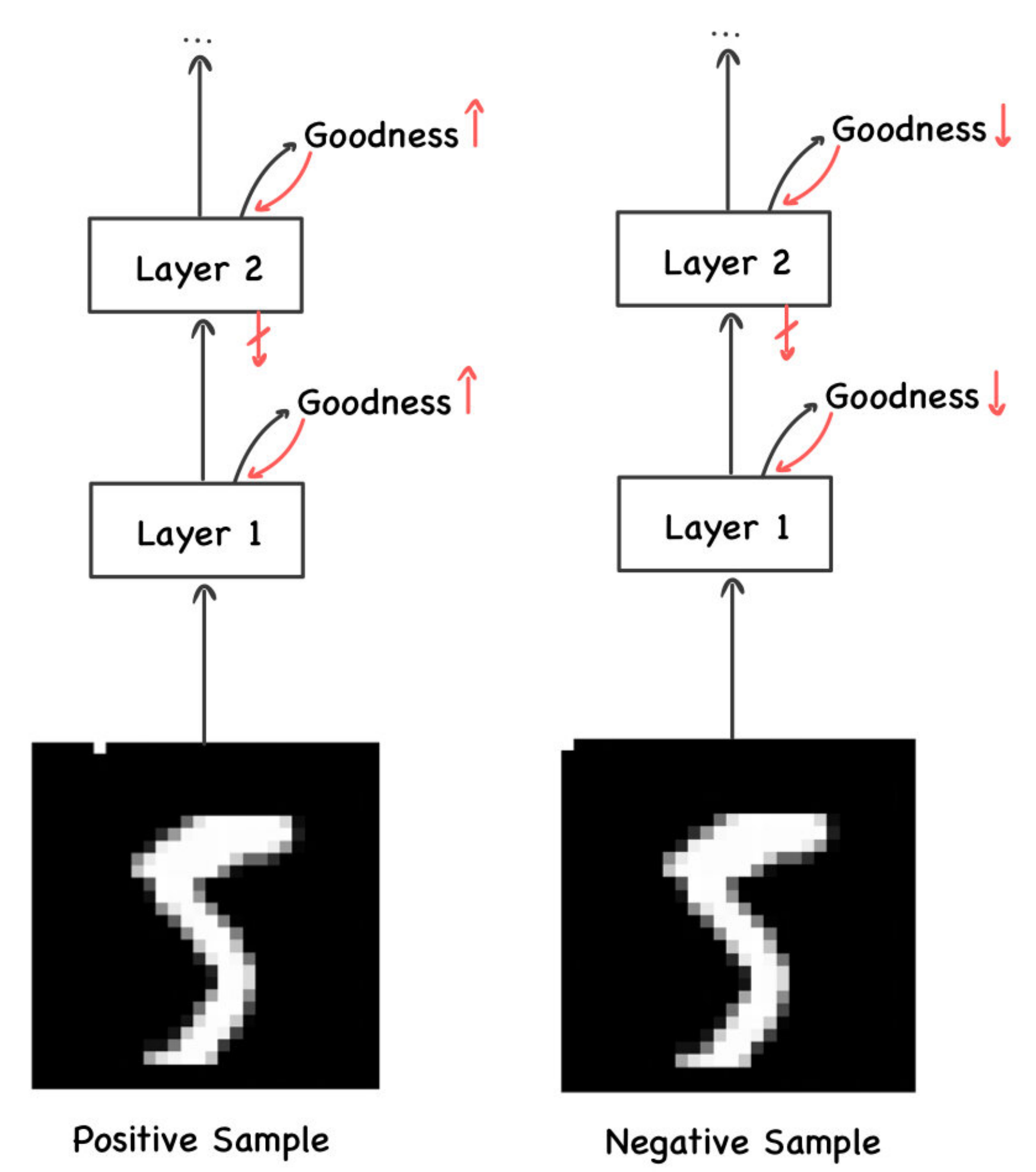}
  \caption{A simplified diagram of a Forward-Forward network. Weights in each layer are adjusted to increase the goodness response of a layer when presented with a positive sample and decrease the goodness response of a layer when presented with a negative sample.~\cite{Lowe}}
  \vspace{-10pt} % space below figure was extending to next page: https://tex.stackexchange.com/questions/111393/too-much-space-around-wrap-figure
\end{wrapfigure}

The Forward-Forward algorithm, proposed and discussed by Hinton~\cite{forwardforward}, is an alternative machine learning method that can replace backpropagation as a learning mechanism within machine learning models. In contrast to backpropagation, which consists of a forward pass used to calculate global objective function followed by backpropagation which adjusts weights based on that global objective function, the Forward-Forward algorithm instead relies on two forward passes and objective functions local to each layer. Optimization of each layer's objective function, also referred to as its "goodness" by Hinton~\cite{forwardforward}, is achieved by adjusting the weights to increase goodness on forward passes with positive data and decrease goodness on passes with negative data. For example, when using a sum of squared activities in the layer as the objective function the algorithm will adjust weights such that the objective function is maximized for positive observations and minimized on negative samples. For an individual layer, accuracy reflects whether a layer exceeded its expected goodness function output for a positive sample, while staying under that threshold for a negative sample. Peer normalization within a layer is employed to prevent extreme activity or inactivity of individual nodes within the layer. For the supervised image classification tasks which are discussed by Hinton~\cite{forwardforward} and the focus of this work, data labels are encoded directly into the black border of the images by replacing one of the first 10 pixels to represent one of the 10 classes. An incorrect pixel is selected for replacement to generate negative data. Motivations in Hinton's work include exploration of more biologically plausible learning methods and support for black-box components within neural networks for which no derivative is known~\cite{forwardforward}.

The motivation of this work is to develop an understanding of the internal behavior and mechanisms underlying Forward-Forward networks by characterizing the behavior of the network as training progresses under a variety of conditions. In particular, this work attempts to answer research questions which may have a fundamental impact on how Forward-Forward neural networks are designed, trained, or analyzed. Section~\ref{sec: future work} discusses potential future work for both developing a greater understanding of Forward-Forward network behavior and possible applications based on theorized general behavior.

Two research questions are explored and supported by a set of experiments that varies the number of training epochs, number of hidden layers, and dimension of each hidden layer. The results of these experiments are analyzed to help explain how Forward-Forward models train by characterizing how layer performance changes based on layer depth and investigating possible correlations between a layer's performance and the overall model. The empirical results reported in this paper suggest that shallower layers may improve sooner than deeper layers and that individual layer accuracy is correlated with overall model accuracy.

\subsection{System, Task, Environment, and Phenomena}

A formulation of the system, task, environment, and phenomena under study are presented to frame the research effort and implemented treatments. For the purposes of this work the system is defined as the computation, or thing carrying out the computation, that produces the phenomena. The task is defined as the objective or purpose of the system. The environment represents the context in which the system does the task. The phenomena represents the computation or behavior that is being studied.

Here, the system includes the Forward-Forward algorithm and its specific properties. Specific system properties include the measure of goodness in each pass, which is currently defined as the sum of the squared neural activities, and peer normalization. The system also includes the specific neural network architecture which employs the Forward-Forward algorithm and is designed to perform particular tasks. Aspects of the architecture which compose the system include the number of layers, dimensions of each layer, number of training epochs, and activation functions used.

This project applies Forward-Forward to image classification tasks, specifically with the MNIST data set~\cite{mnist}. The task also includes the method used to encode data labels and to generate negative data for the second forward pass. In this case, data labels are encoded directly into the black border of the MNIST images by replacing one of the first 10 pixels to represent one of the 10 classes. An incorrect pixel is selected for replacement to generate negative data. Alternative methods of generating negative samples or testing with different data sources, like CIFAR-10~\cite{cifar}, are also possible.

The environment includes, arguably, the data set chosen, the programming language and libraries used, and the platform on which the model is deployed and executed. This project employs MNIST, a PyTorch implementation, and an ARM64 architecture Apple Macintosh platform.

The primary phenomenon studied is performance, represented by the accuracy and loss, for each individual layer in the network. This phenomenon may also be interpreted as the composition of two, or more, distinct phenomena - those occurring within individual layers versus that of the overall neural network. The goal of studying this phenomenon is to better understand and explain how individual layer performance impacts overall model performance, which may in turn influence how these types of models are engineered.

\subsection{Related Work}

A common motivation amongst researchers~\cite{forwardforward, end_to_end, local-error-signals} investigating novel training mechanisms as alternatives to backpropagation is the pursuit of learning approaches and theories which are more biologically plausible. Most current approaches employ backpropagation and stochastic gradient descent, which while effective, learn in ways anomalous compared to our current understanding of biological systems~\cite{forwardforward}. Related work often places much consideration on locality when proposing new solutions, favoring constrained local approaches rather than global objective functions. Researchers have explored exploiting locality in the form of sub-networks with heterogeneous training approaches~\cite{random-variations}, sub-networks with homogeneous training approaches~\cite{end_to_end}, and training by individual layer~\cite{forwardforward}. Hybrid architectures have also been explored in related work~\cite{random-variations} which incorporate Forward-Forward and backpropagation sub-networks to achieve greater performance or the flexibility to include non-differential components within a network. A common theme among these approaches is that training is local: subsequent layers or networks have no impact on preceding layers and local loss functions do not depend on global error.

As "Preliminary Investigations..." in the title of Hinton's paper suggests, investigation of Forward-Forward networks is still relatively nascent~\cite{forwardforward}. Research into large scale Forward-Forward models is lacking and, similarly, applications are focused on classification rather than regression or generative applications. Specifically, image classification dominates the reviewed literature, although audio classification~\cite{end_to_end} receives some treatment. In general, much of the current research reviewed focuses on novel techniques and improvements in overall performance~\cite{end_to_end, random-variations}. Research which explores characteristics of internal model performance based on existing methods, like the internal behavior of Forward-Forward networks, receives less treatment, but could provide crucial explanatory value which could influence future research.

\section{Research Questions and Hypotheses}
% 1. Choose research question(s) and the hypotheses. Select the research questions and hypotheses that you intend to address. In general, you should select only a single research question and the hypotheses related to it. You can always add additional questions later, particularly if your first choice doesn’t produce interesting results. However, attempting to analyze several research questions may leave you with little time to examine any of them thoroughly. It is better to answer a one research question well than to answer several questions badly.

Two research questions are proposed and supported by data collected from a common set of experiments, which allows analysis of multiple potential interesting phenomena without additional empirical effort. The research questions are focused on characterizing the behavior of Forward-Forward networks in order to better explain and understand their operation, rather than on producing state-of-the-art performance or proposing a new approach.

\begin{center}
\fbox{\begin{minipage}{0.8\textwidth}
\textbf{Research Question 1:} How does the depth of layer within a Forward-Forward network affect its training behavior?
\end{minipage}}
\end{center}

Research Question 1 explores how Forward-Forward networks train. Unlike backpropagation, where layer activation adjustments are made to optimize for a downstream loss function, Forward-Forward layers operate on a local loss function. Thus, with a feed forward architecture, layer training is unaffected by subsequent layers when using the Forward-Forward algorithm and may behave differently than models employing more traditional learning approaches do. Three hypotheses are proposed based on Research Question 1:

\textbf{Hypothesis 1:} \emph{Layer goodness improvement in deeper layers is delayed as improvement cascades through the network.}

\textbf{Hypothesis 2:} \emph{All layers improve simultaneously.}

\textbf{Hypothesis 3:} \emph{Layer goodness improves sooner in deeper layers than in shallower layers.}

These hypotheses theorize how loss improvement evolves in different layers based on their depth in the network. Hypothesis 1 would be falsified if no delay in achieving the target layer accuracy is observed; conversely, Hypothesis 2 would be falsified if a delay is observed. Hypothesis 3 would be falsified if shallow layers achieve a target value faster, i.e., with less delay, than deeper layers. For the purposes of these hypotheses, the delay is measured by the epoch at which a specific layer achieves or exceeds a target accuracy. 

\begin{center}
\fbox{\begin{minipage}{0.8\textwidth}
\textbf{Research Question 2:} How strongly does a layer's goodness correlate with overall classification accuracy?
\end{minipage}}
\end{center}

Research Question 2 investigates whether an individual layer's accuracy telegraphs final overall classification accuracy. The answer to this research question could have practical applications to the construction of Forward-Forward networks or influence how the performance of Forward-Forward networks is analyzed.

\textbf{Hypothesis 1:} \emph{An individual layer's goodness strongly correlates with overall classification accuracy.}

\textbf{Hypothesis 2:} \emph{An individual layer's goodness does not strongly correlate with overall classification accuracy.}

\textbf{Hypothesis 3:} \emph{Layers deeper in the network exhibit stronger correlation with overall classification accuracy than shallower layers.}

Hypothesis 1 predicts a strong correlation between layer accuracy and overall classification accuracy. This hypothesis is important as, if true, it may be possible to incrementally train and construct Feed-Forward networks such that additional layers are only added when required. Hypothesis 2 represents an alternative possibility where no useful correlation is discovered between individual layer accuracy and overall classification accuracy. For Hypothesis 1 and 2, a strong correlation is defined as a Spearman or Pearson correlation coefficient greater than 0.7. Hypotheses 3 explores how strength of correlation may vary based on the depth of the layer in the network and would be falsified if no difference in correlation is discovered based on layer depth or if shallower layer accuracy exhibited stronger correlation with overall model accuracy than deeper layers.

\section{Research Design}

The default system configuration is composed of 4 hidden layers, each of which contains 1000 ReLU activation functions. The primary treatment involves variation in the number of hidden layers comprising the system. The number of hidden layers tested is $\{ 2, 4, 8, 16, 32 \}$. Variations in hidden dimensions per layer are also performed to allow collection of results under a wider variety of conditions and to investigate potential impacts of the number of activations in a layer. Lastly, variation in the number of training epochs is considered by conducting measurements at the conclusion of each epoch during training. Execution of each experimental configuration is repeated 30 times with different random number generator initialization seeds. In total, there are 10 unique experimental configurations, not including seeds, which are repeated 30 times for a total of 300 distinct iterations of model training. Table~\ref{table: exp2} presents a fuller view of the experimental configuration.

\begin{table}[h]
  \caption{Experimental configuration. The full suite of experimental treatments is represented by the cartesian product of the seed, hidden dimensions per layer, and number of layers denoted in the table. }
  \label{table: exp2}
  \centering
  \begin{tabular}{lll}
    \toprule
    Name     & Value \\
    \midrule
    Seed                        & $\{$ 1, 2, 3, ... , 30 $\}$ \\
    Device                      & "mps"     \\
    Batch Size                  & 1000     \\
    Peer Normalization          & 0.03      \\
    Peer Normalization Momentum & 0.09      \\
    Hidden Dimensions per Layer & $\{$ 100, 1000 $\}$      \\
    Number of Layers            & $\{$ 2, 4, 8, 16, 32 $\}$         \\
    Epochs                      & 30        \\
    Learning Rate               & 1e-3      \\
    Weight Decay                & 3e-4      \\
    Momentum                    & 0.9       \\
    Downstream Learning Rate    & 1e-2      \\
    Downstream Weight Decay     & 3e-3      \\
    Validation Index            & 1         \\
    \bottomrule
  \end{tabular}
\end{table}

The objective function employed by each hidden layer is the sum of squared activities of all activation functions in a layer. The objective is to maximize this function for positive observations, while minimizing it for negative sample. Given the activation $a_j$ of activation function $j$ within a layer and a threshold $\theta$ the objective function is represented by $y$~\cite{forwardforward}.

\begin{equation}
\label{eq: objective}
y = \sigma \left( \sum_{j}a_j^2 - \theta \right)
\end{equation}

This objective function is optimized via maximization or minimization based on the type of sample. The function $J$ represents the piecewise objective function which defines the optimization of the objective function over the domain of sample types and where $x^i$ represents the $i^{th}$ sample.

\begin{equation}
\label{eq: piecewise}
  J =
  \begin{cases}
                                   \max y & \text{if $x^i$ is positive sample} \\
                                   \min y & \text{if $x^i$ is negative sample} \\
  \end{cases}
\end{equation}

Individual layer accuracy is defined as the sum of true positives, $TP$, and true negatives, $TN$, over the number of samples, $m$. For each layer, a response is considered a true positive if the sample is positive and the objective function exceeded the threshold. Similarly, a true negative occurs when a sample is negative and the objective function does not exceed the threshold.

\begin{equation}
\begin{aligned}
\label{eq: accuracy}
& TP = \sum_{i}[(y^i > 0.5) \land (\text{$x^i$ is positive sample})] \\
& TN = \sum_{i}[(y^i < 0.5) \land (\text{$x^i$ is negative sample})] \\
& Accuracy = \frac{TP + TN}{m}
\end{aligned}
\end{equation}

In pursuit of Research Question 1, overall model classification accuracy and individual layer accuracy is computed after each epoch. The figures presented are largely based on the average accuracy across all repetitions based on initialization seed; Figure~\ref{fig:error_bars} visualizes the deviation in accuracy across repetitions. The delay is measured by the epoch in which a specific layer achieves or exceeds a target accuracy. Figure~\ref{fig:delay_subplot} provides a visualization of the computed delay based on layer depth for a target accuracy of 0.7.

For Research Question 2, correlation strength between layer goodness and overall classification accuracy is determined by the absolute value of Pearson's correlation coefficient and Spearman's correlation coefficient where a value greater than 0.7 indicates strong correlation. Pearson's correlation coefficient is utilized because it measures the linear relationship between two sets of data~\cite{Rovetta}. The Spearman's rank coefficient is also be calculated as a measure of any monotonic relationship, i.e., including non-linear relationships~\cite{Rovetta}. The correlation between a specific layer and overall model output is based on layer depth with each correlation coefficient value based on 900 pairs of data points which are collected across all 30 repetitions and at each epoch over 30 epochs of training. Figure~\ref{fig:empirical_1_corr} visualizes these correlation coefficients for each experimental configuration as a function of layer depth.

\section{Experimental Results}

\begin{figure}
         \centering
         \includegraphics[width=\textwidth]{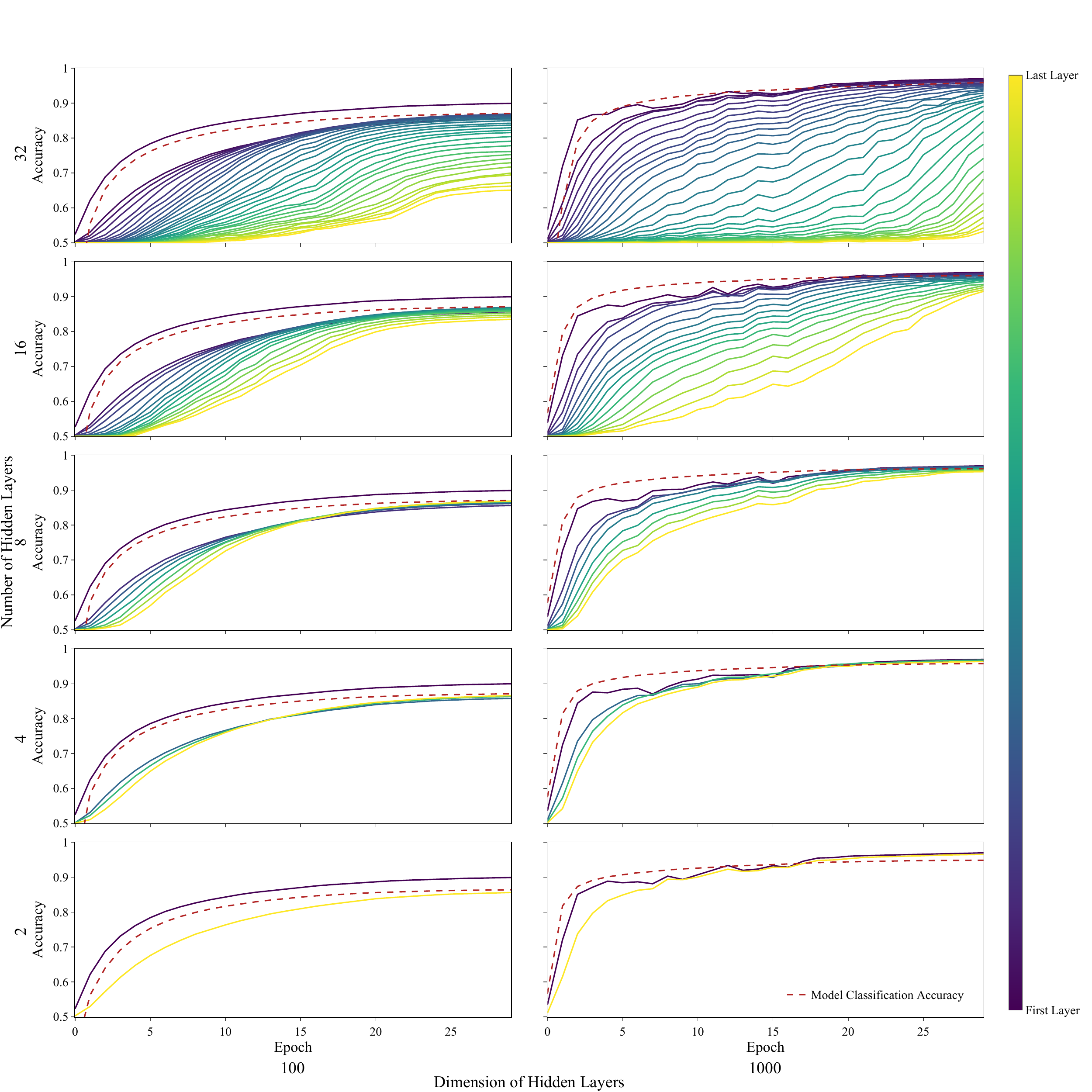}
         \caption{Layer accuracy as training progresses. Each subplot represents a different experimental configuration; subplot column is defined by the dimension of each individual hidden layer while subplot row indicates the number of hidden layers. For each individual subplot the x-axis represents the training epoch at which measurement occurred and the y-axis represents the accuracy measured. Each data point represents the average accuracy across 10 repetitions with different initialization seeds. Trace color represents relative layer depth. Overall model accuracy is indicated as a dashed line}
         \label{fig:accuracy_subplot}
\end{figure}

\begin{figure}
         \centering
         \includegraphics[width=\textwidth]{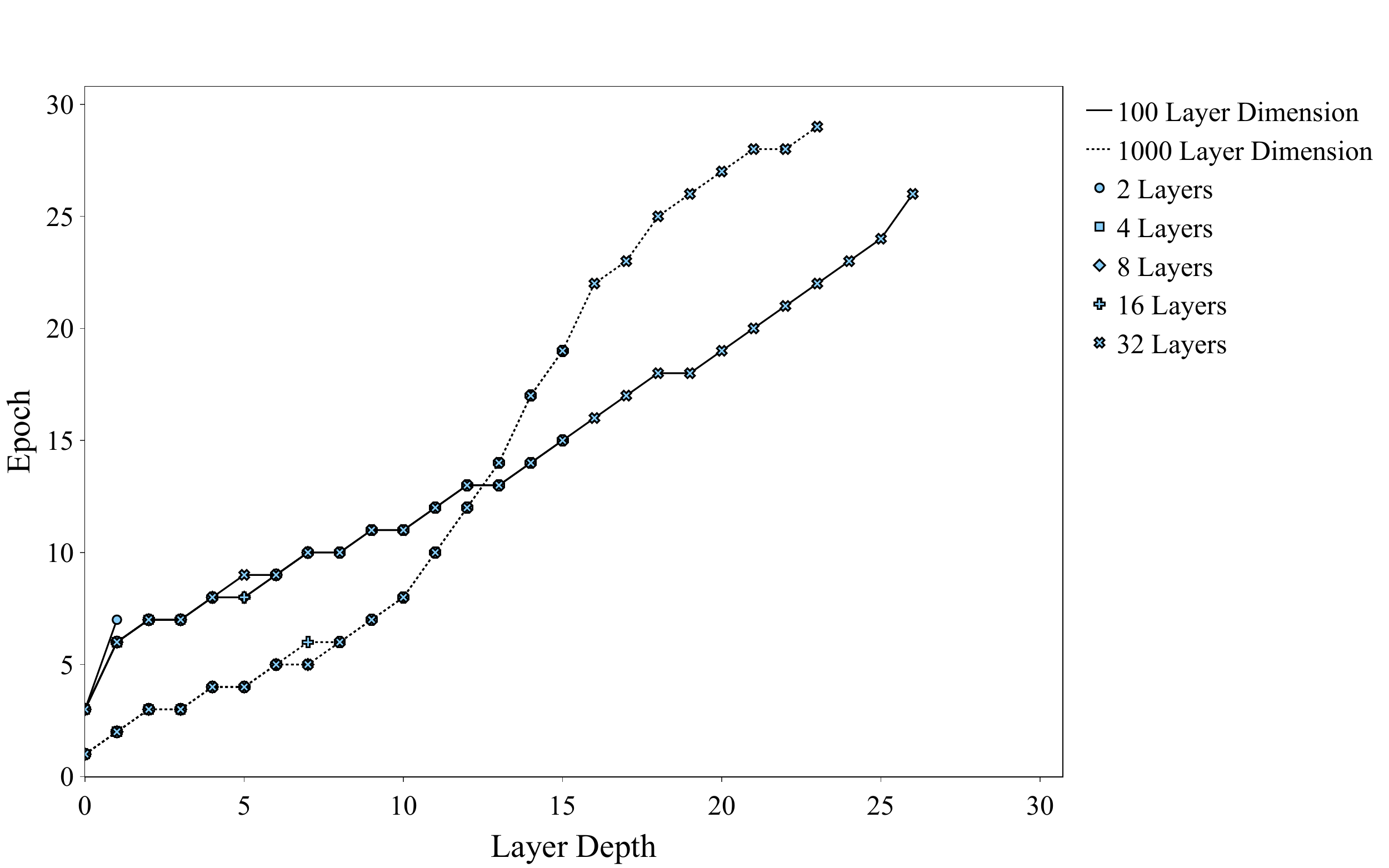}
         \caption{Epoch at which a layer achieves individual target accuracy of 0.7. Each trace represents a specific model configuration defined by the number of hidden layers, denoted by marker style, and the dimension of each hidden layer, denoted by line style. The points on each trace are determined based on the epoch at which a layer at a specific depth achieves the target layer accuracy of 0.7. The x-axis indicates the layer depth, while the y-axis indicates the epoch at which the target was reached. In relation to Figure~\ref{fig:accuracy_subplot}, this figure represents the epoch value of intersection points of a horizontal line at 0.7 accuracy with each individual layer accuracy trace.}
         \label{fig:delay_subplot}
\end{figure}

Figure~\ref{fig:accuracy_subplot} depicts variations in individual layer accuracy as training progresses. In general, these results support Hypothesis 1: that layer goodness improvement, represented here by accuracy, is delayed in deeper layers. For that same reason it appears that Hypothesis 2, which predicts no delay in improvement, is false. The delay effect appears to be more pronounced as the number of hidden layers increases, but this is likely due to the number of plot traces which assist in visualizing the effect, rather than a change in the effect itself, based on analysis of Figure~\ref{fig:delay_subplot} which indicates that layers at specific depths perform similarly regardless of the number of hidden layers.

\begin{figure}
         \centering
         \includegraphics[width=\textwidth]{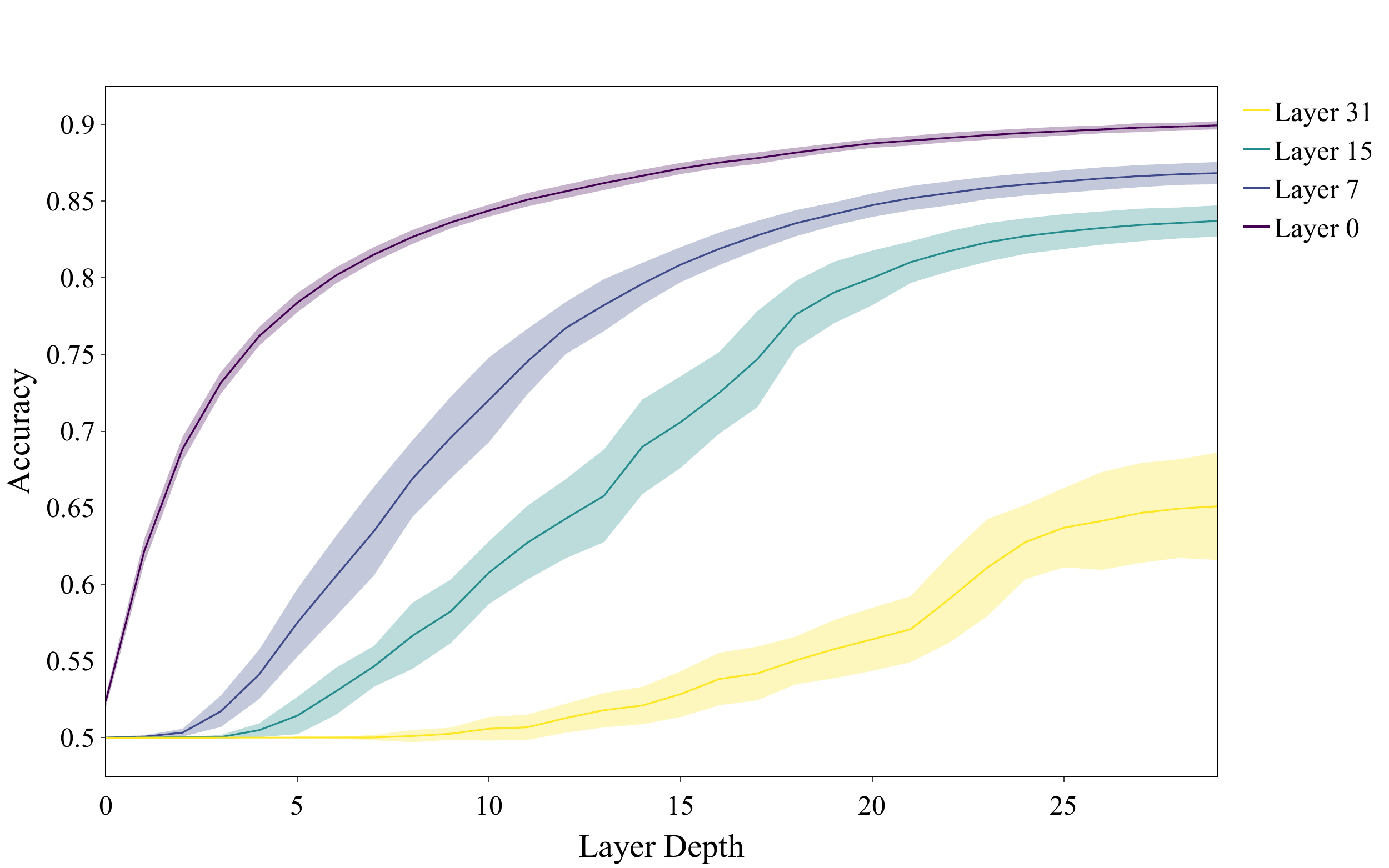}
         \caption{Continuous error bars for a selection of layers for a Forward-Forward model with 32 layers and 1000 activations per layer. The error bars represent one standard deviation from the mean based on 30 repetitions with random initialization seeds.}
         \label{fig:error_bars}
\end{figure}

Figure~\ref{fig:error_bars} depicts the measured standard deviation for a single configuration and a subset of layers to demonstrate how deviation in empirical results may affect the strength of any inferred conclusions. In general, deviation in measured accuracy tends to be larger in deeper rather than shallower layers. The experimental deviation is narrow enough to suggest that the general delay behavior examined is largely consistent as no significant overlap across layers exists across a single standard deviation.

Figure~\ref{fig:delay_subplot} provides a visualization of the delay, where delay is measured by the epoch at which a specific layer accuracy is achieved. The slope of each trace in Figure~\ref{fig:delay_subplot} can infer the behavioral characteristics of the delay. For example, a horizontal trace would indicate that all layers achieve the target layer accuracy at the same time, while a positive slope would indicate that deeper layers achieve the target accuracy at later epochs than shallower layers. Figure~\ref{fig:delay_subplot} illustrates a distinct difference in behavior based on hidden layer dimension; shallow layers in models with hidden layers composed of 1000 activation functions per layer consistently achieve the target accuracy faster than models containing 100 activation functions per layer. In contrast, deeper layers in models with hidden layers of 1000 activation functions achieve the target accuracy at later epochs than the narrower models with hidden layers containing 100 activation functions. Notably, for a fixed layer dimension, a layer at a specific layer depth experiences the same delay regardless of the total number of hidden layers in the network. Additionally, for the configurations tested and in particular for configurations which employ 100 dimension layers, the change in delay across network depths appears largely constant.

Based on the analysis of the delay observed in Figure ~\ref{fig:accuracy_subplot} and Figure ~\ref{fig:delay_subplot}, Hypothesis 1 and Hypothesis 3 of Research Question 1 appear false: individual layers do not improve simultaneously nor do deeper layers achieve the target accuracy more quickly than shallow layers.

\begin{figure}
     %\centering
     %\captionsetup{justification=centering}
     \begin{subfigure}[b]{0.45\textwidth}
         \centering
         \includegraphics[width=\textwidth]{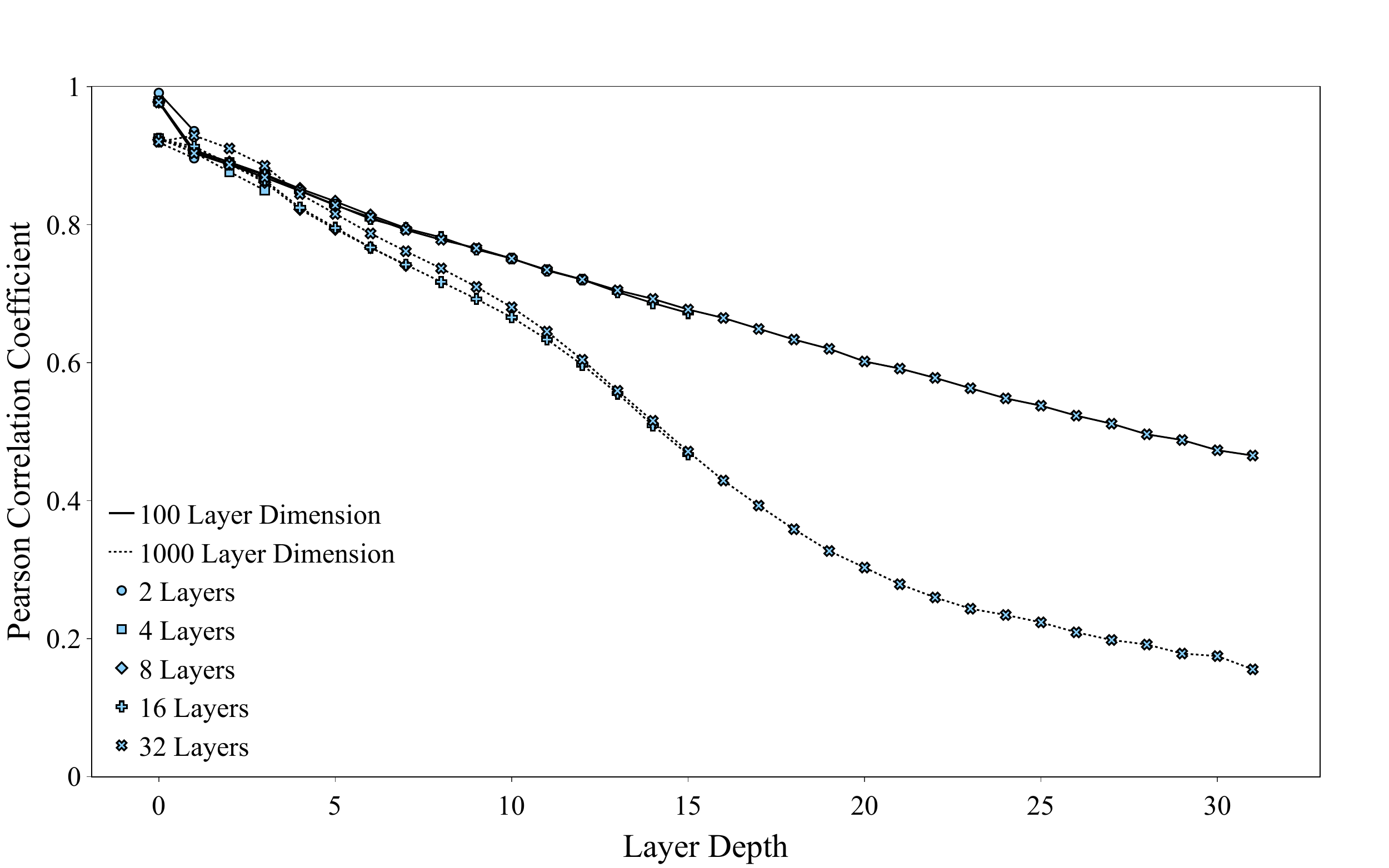}
         \caption{Pearson Correlation}
         \label{fig:pearson}
     \end{subfigure}
     \hfill
     \begin{subfigure}[b]{0.45\textwidth}
         \centering
         \includegraphics[width=\textwidth]{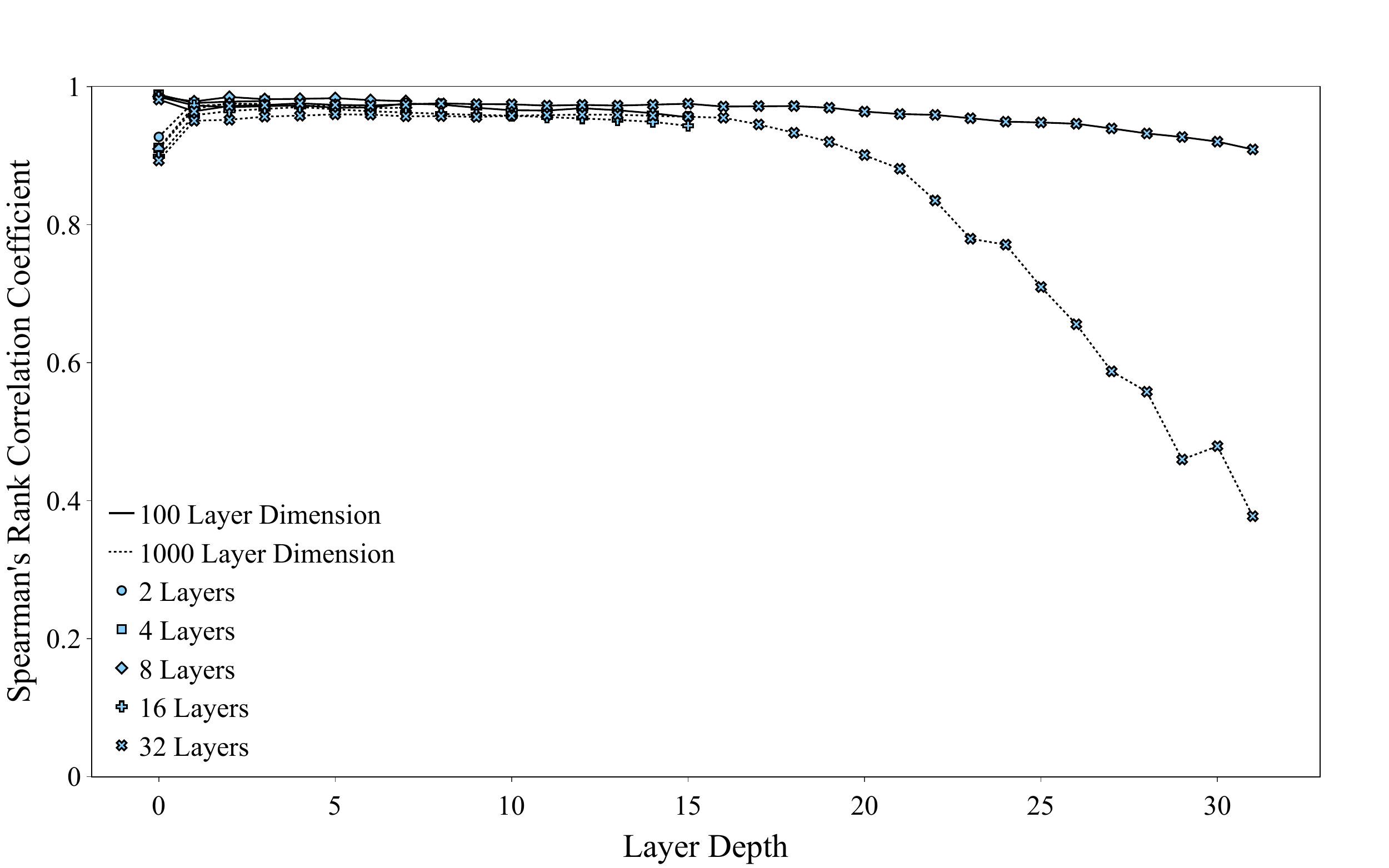}
         \caption{Spearman's Rank Correlation}
         \label{fig:spearman}
     \end{subfigure}
     \caption{Correlation of individual layer accuracy with overall network accuracy.~\ref{fig:pearson} illustrates the Pearson correlation coefficient at each layer and provides a measure of any linear relationship between individual layer accuracy and overall network accuracy.~\ref{fig:spearman} illustrates Spearman's rank correlation coefficient at each layer and provides a measure of any monotonic relationship between individual layer accuracy and overall network accuracy. Both Pearson and Spearman's rank correlation coefficient may vary between -1 and 1; the sign of the coefficient represents the direction of the relationship while the magnitude of the coefficient represents the strength of the relationship. The correlation for each layer is based on all accuracy measurements across every epoch for the layer and for each experimental repetition with varied initialization seeds.}
    \label{fig:empirical_1_corr}
\end{figure}

Figure ~\ref{fig:empirical_1_corr} provides plots for both the Pearson and Spearman's rank correlation coefficients which measure the correlation between individual layer accuracy and overall network accuracy. In general, layer accuracy appears to strongly correlate with overall network accuracy regardless of depth, although the linear correlation indicated by the Pearson correlation coefficient significantly weakens as layer depth increases. Increase in layer width also appears to significantly weaken the observed correlation between layer and overall model accuracy with particular acuity at deeper layers. Qualitative inspection of the overall accuracy relative to layer accuracy in Figure ~\ref{fig:accuracy_subplot} strengthens the finding that shallower layers more strongly correlate with overall model accuracy than do deeper layers as in all subplots overall model accuracy more closely aligns with shallower rather than deeper layers.

The results presented indicate a more complex relationship between layer accuracy and overall model accuracy than the proposed hypotheses would suggest. Both Hypothesis 1 and 2 of Research Question 2 are false as whether or not an individual layer strongly correlates with overall model accuracy is conditioned, at least, on layer depth or some other implicit confounding variable.

\section{Threats to Validity}

These experiments may not generalize as only a very limited subset of system, task, and environment treatments were applied. Larger variations in number of training epochs and layer dimensions are needed to more fully explore their impact on the observed behavior. Experimentation on untreated system characteristics including learning rate, batch size, and peer normalization, is also needed to root out possible confounding variables which prevent causal relationships from being proposed with any confidence.

The current experimental approach, which focuses on treatments to the system, poses threats to external validity. These experiments focus solely on image classification tasks with the MNIST data set, which limits confidence in these conclusions to those applications in the absence of further testing. Further testing on other tasks, including regression or generation, and on other datasets is necessary to provide greater confidence in the generalizability of results presentd.

Internal threats to validity include the method of determining correlation and the limited number of experimental conditions evaluated. For example, correlation is currently measured based on pairs of data from across all epochs, but the strength of direction of any relationship may be different depending on the number of epochs or span of epochs evaluated. Additionally, the influence of overall model accuracy on correlation strength is not evaluated and may be a confounding variable responsible for any correlation differences observed in models of various size. As with all experiments, additional repetitions across a greater variety of conditions would also help strengthen validity.

\section{Future Work}
\label{sec: future work}

This work only considers image classification tasks on a small dataset. To corroborate these results, future work may include experimentation on other datasets, such as CIFAR10~\cite{cifar} or Pets~\cite{pets}, within the image classification domain to explore any phenomena which may be attributed to the dataset chosen rather than the model itself. There is also significant opportunity to explore variations in system characteristics as well. Much is left unexplored in this research, including the effect of variation in activation function, peer normalization, and learning rates. Novel layer objective functions, other than the sum of squared activities for each layer, may also have a significant impact on model behavior.

Analytically, future work may also consider evaluation of layer logits directly, rather than predicted class or objective function result alone. Exploration of logits directly may unveil new phenomena or explanatory detail into model behavior and performance. In particular, analysis of raw logits may provide insight into the relationship between deep layers and overall model performance. Researchers may also consider exploring the delay or correlation in relation to other measured attributes besides layer depth; for example, overall model accuracy may influence how these phenomena vary.

Additional efforts may also consider the application of theories of Forward-Forward network learning mechanism to the engineering of those networks. As layers in a feedforward Forward-Forward network only rely on preceding layers and inputs, rather than subsequent layers in the case of backpropogation, it may be possible to iteratively construct Forward-Forward networks by only adding additional layers as necessary. A methodological approach to iterative construction of these networks may be possible based on the understanding of both accuracy propagation through layers and the correlation of a layers performance with overall model accuracy.

\section{Conclusion}

The Forward-Forward algorithm is a promising new machine learning approach that provides an alternative to backpropagation. As Forward-Forward networks operate in very different ways than backpropagation-based networks, further research into phenomena related to internal model performance is necessary to fully exploit the approach when designing and training these networks.

This paper presents an investigation into the training behavior of Forward-Forward networks, with a focus on how layer accuracy changes based on the depth of the layer and system hyperparameters. The results suggest that deeper layers achieve target accuracy values at later epochs than shallower layers and that shallower layers may more strongly correlate with overall model accuracy than deeper layers.

These results represent only a small fraction of possible configurations and contexts for Forward-Forward networks. This initial investigation explores intriguing phenomena related to the behavior of Forward-Forward networks which may both strengthen our understanding of these networks and how to apply them effectively.

\FloatBarrier

%\clearpage % Ensure all figures are placed before adding references
{
\small

\bibliographystyle{plain}
\bibliography{refs}
}

%\newpage

% \section{Supplementary Material}

% \FloatBarrier

% \begin{table}[ht]
%   \caption{Layer and Model Accuracy Standard Deviation. This table presents the maximum standard deviation in accuracy observed with various system configurations.}
%   \label{table: exp2}
%   \centering
%   \begin{tabular}{lllll}
%     \toprule
%     \multirow{3}{*}{Number of Hidden Layers} &
%     \multicolumn{4}{c}{Dimension of Hidden Layers} \\
%     \cmidrule(lr){2-5}
%     & \multicolumn{2}{c}{Layer} & \multicolumn{2}{c}{ Model } \\
%     \cmidrule(lr){2-3} \cmidrule(lr){4-5}
%     & 100 & 1000 & 100 & 1000\\
%     \hline
%     32 & 0.035 & 0.032 & 0.027 & 0.071\\
%     \hline
%     16 & 0.036 & 0.070 & 0.024 & 0.039\\
%     \hline
%     8 & 0.028 & 0.038 & 0.023 & 0.033\\
%     \hline
%     4 & 0.0205 & 0.031 & 0.0270 & 0.032\\
%     \hline
%     2 & 0.009 & 0.0325 & 0.0308 & 0.0386\\
%     \hline
%   \end{tabular}
% \end{table}

\end{document}